\pdfoutput=1 % force arXiv to build with pdflatex (figures are PDF/PNG; the legacy latex/DVI path would fail)
\documentclass[letterpaper, 10pt, journal, twoside]{IEEEtran}

\usepackage{graphics} %
\usepackage{epsfig} %
\usepackage{mathptmx} %
\usepackage{times} %
\usepackage{amsmath} %
\usepackage{amssymb}  %
\usepackage{siunitx} 
\usepackage{soul}
\usepackage{cite}
\usepackage{bm}
\usepackage{booktabs}
\usepackage{microtype}
\usepackage{graphicx}
\usepackage{subcaption}
\usepackage{flushend}
\usepackage{etoolbox}
\usepackage{algorithm}
\usepackage{algorithmic}
\usepackage{marginnote}
\usepackage{xcolor}

\usepackage{enumitem}

\newif\ifdraft
\draftfalse

\ifdraft
  \usepackage[textsize=tiny]{todonotes}
  \let\oldmarginpar\marginpar
  \renewcommand{\marginpar}[1]{\oldmarginpar{\footnotesize\color{blue}#1}}
\else
  \usepackage[disable]{todonotes}
  \renewcommand{\marginpar}[1]{}
  
\fi

\newif\ifrevision
\revisionfalse   %
\newcommand{\rev}[1]{\ifrevision\textcolor{blue}{#1}\else#1\fi}

\usepackage{caption}
\usepackage{etoolbox}
\AtBeginEnvironment{algorithmic}{\footnotesize}

\title{LOPAL: Local Performance-Aware Active Learning\\
from Imperfect Demonstrations}

\author{Johannes Heidersberger$^{1}$, Shail Jadav$^{1}$, and Dongheui Lee$^{1,2}$%
\thanks{This work was supported by the Austrian Science Fund (FWF) under the project ``LunarAssembly'' [PIN5145724], and by the Vienna Science and Technology Fund (WWTF) under the project ``SafeDiffusion'' [ICT25068]. The authors acknowledge TU Wien Bibliothek for financial support through its Open Access Funding Program. }

\thanks{$^{1}$Johannes Heidersberger, Shail Jadav, and Dongheui Lee are with Autonomous Systems Lab, Institute of Computer Technology, TU Wien, Vienna, Austria
        {\tt\footnotesize \{johannes.heidersberger, shail.jadav, dongheui.lee\}@tuwien.ac.at}}%
\thanks{$^{2}$Dongheui Lee is with Institute of Robotics and Mechatronics, German Aerospace Center (DLR), Weßling, Germany.}%
\thanks{This article has been accepted for publication in IEEE Robotics and Automation Letters. \copyright~2026 The Authors. This work is licensed under a Creative Commons Attribution 4.0 International (CC BY 4.0) license. Digital Object Identifier (DOI): 10.1109/LRA.2026.3698364}%
}
\begin{document}

\maketitle

\begin{abstract}
    Learning from Demonstration (LfD) enables intuitive robot skill acquisition by allowing robots to learn directly from human task demonstrations. However, current methods often fail to address the fact that due to suboptimal and inconsistent human behavior, the quality of the demonstration
    can vary within each demonstration. Therefore, we introduce LOPAL (LOcal Performance-aware Active Learning), an active learning approach that leverages this local demonstration quality information. Our approach consists of two synergistic components. First, a local performance-driven LfD method uses a Gaussian Mixture Model (GMM) to encode both the demonstrated trajectories and their associated local quality assessments. This enables the generation of trajectories that outperform the imperfect demonstrations by utilizing complementary local data of high performance. Second, active data acquisition allows to improve beyond the imperfect demonstrations by collecting additional informative samples.
    In areas missing good data, the user is actively requested to provide corrections through a shared autonomy (SA) mechanism, while the robot autonomously executes the learned behavior.
    The efficacy of LOPAL was validated in both a simulation and a real-world experiment. 
    The results from a real-world pipe inspection task showed that the proposed approach can achieve up to $27.31\%$ improvement in task performance while also reducing the effort required to collect the demonstrations.
\end{abstract}


\section{INTRODUCTION}
\IEEEPARstart{T}{raditional} manual programming can be time-consuming and laborious.
Learning from Demonstration (LfD) has emerged as a promising alternative, allowing users to intuitively teach new skills to robots through human demonstrations \cite{lee2011incremental}.
Despite these advantages, a fundamental challenge in LfD is the strong dependence on the quality of the demonstrations. 
In this work, we consider demonstration quality in terms of task performance, i.e., the degree to which the demonstrated behavior satisfies the task specifications.
If the robot naively replicates the suboptimal demonstrations, it risks reproducing the demonstrated flaws. 
This challenge is compounded by the inherent variability in human performance during the demonstration. Even a demonstration that is of overall high quality may contain segments of suboptimal behavior \cite{yu2022increased}. Human task performance is contingent not only upon skill level but also upon extraneous factors such as fatigue\rev{\cite{sam_impact_2024}}.
For tasks involving extended operational periods, it is common to observe a decrement in individual performance as the task progresses.
Active Learning (AL) allows to reduce the demonstration burden on the user by enabling the robot to actively collect the additional data that it needs to improve its performance.
This research endeavors to advance active learning from demonstration by explicitly accounting for \rev{local performance, i.e.,} local variations of task performance in human demonstrations.

\subsection{Related Works}
\begin{figure} [t]
    \centering
    \includegraphics[width=0.95\linewidth]{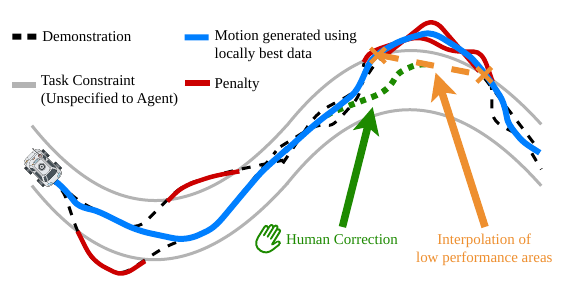}
    \caption{Conceptual illustration of the proposed LOPAL framework for learning from imperfect demonstrations. The demonstrations (black dashed) contain local imperfections, such as penalties (red) when a vehicle leaves the track (gray).
    Leveraging the local quality information \rev{derived from these penalties} allows to generate a motion (blue) that follows the locally best available motion.
    If all demonstrations are of low quality, the motion is interpolated between adjacent good data points (orange) to estimate a better nominal behavior. During execution, the user is then actively requested to provide better demonstration by correcting this nominal behavior (green).}
    \label{fig:hrisetup}
    \vspace{-0.0cm}
\end{figure}
LfD research has begun to recognize the imperfect nature of human demonstrations, yet intra-demonstration variability of quality is still largely overlooked.
Early work introduced the concept of learning from failed demonstrations
\cite{grollman_donut_2011}, which was later extended to incorporate both successful and unsuccessful examples \cite{hertel_learning_2021}.
Such a binary classification provides a clear signal for a learning algorithm to differentiate between desired and undesired behaviors.
Other methods use continuous score values to rate demonstration quality
\cite{cao_learning_2021, hoang_sprinql_2024},
enabling a more granular prioritization of higher-quality demonstration data during the learning process.
In some works, explicit local quality measures are computed directly from the demonstration data 
\cite{bilal2024beyond, sakr_consistency_2025}.
However, these local measures are then collapsed into global scores used to weight the demonstration data for learning.
Other work indirectly estimates demonstration quality at a local level from global performance assessments 
\cite{yin_offline_2024}, which allows robots to selectively learn from local high-quality data.
By relying on global quality ratings and local ratings derived from them, current LfD methods obscure local quality variations within demonstrations.
When two demonstrations contain low-quality data in different regions, relying solely on their global assessments cannot reveal which local behaviors are superior.
Consequently, their capability to identify and integrate high-quality sub-trajectories into a potentially superior overall behavior is limited.
\\
Certain imperfections of demonstrations have also been handled without explicitly assessing the demonstration quality. In particular, missing data and large temporal or spatial variations between demonstrations were addressed by aligning the demonstrations containing these imperfections to a reference demonstration \cite{pervez_novel_2017, pervez_motion_2019}.
\\
Besides pure imitation learning, another prominent direction for improving upon imperfect demonstrations is to combine LfD with Reinforcement Learning (RL).
In this paradigm, demonstrations are typically used to bootstrap a Reinforcement Learning agent, for instance by pre-training a policy, which is then fine-tuned through exploration to discover behaviors superior to the ones demonstrated\rev{\cite{ren_diffusion_2025, ball_efficient_2023}}.
However, this fine-tuning often requires extensive exploration, which can be time-consuming and \rev{in real world scenarios} potentially unsafe.
\rev{The exploration can be guided by human critique of generated trajectories to refine a learned reward \cite{hirota_active_2025}, shifting the human teaching effort from demonstrations to evaluations of the robot behavior.}
\\
In addition to active exploration, which can be performed autonomously by the robot, requesting additional user demonstrations is another approach for obtaining additional data in AL.
While user-initiated interventions can be used to demonstrate better behavior \cite{luo_precise_2025}, actively requesting corrections can reduce the user burden by focusing the human input where it is most needed.
This active querying of additional user demonstrations within AL frequently involves dynamically distributing the control between the human and a robot agent (SA)
\cite{michel_learning-based_2023, jadav_shared_2024}.
The robot can actively indicate the need for user input, for example by adjusting the speed \cite{eiband_collaborative_2023}, modulating the compliance, or providing visual feedback \cite{celemin_knowledge-_2023}.
Multiple active imitation learning approaches request demonstrations based on aleatoric and epistemic uncertainty measures 
\cite{celemin_knowledge-_2023, eiband_collaborative_2023}.
However, these measures reflect ambiguity over which action to take, not necessarily uncertainty about which demonstrated action may lead to the best task outcome.
To the best of our knowledge, existing AL methods have not yet incorporated local variations in demonstration quality to guide data acquisition toward regions where it would most enhance the robot’s performance.
As a result, data may be over-collected in well-performing regions and under-collected in more difficult ones.

\subsection{Contributions}
In this work, we propose LOPAL (LOcal Performance-aware Active Learning), an AL framework illustrated in Fig. \ref{fig:hrisetup}, which leverages local performance information to guide both learning from demonstration and active data acquisition, with the aim of enhancing task performance and reducing user effort.
Unlike existing LfD methods, which do not consider local quality variations, our approach explicitly leverages local performance information to combine the best parts of multiple imperfect demonstrations, creating a behavior that outperforms any single demonstration.
Furthermore, data collection is actively targeted at regions where high-quality demonstration data is missing, thereby aiming to reduce the user's teaching effort.
Our contributions are:
\begin{enumerate}
    \item A local performance-aware LfD method that aims to generate trajectories replicating the best demonstrated local behavior.
    \item An active data acquisition approach, combining autonomous performance-guided local motion modulation with human input through a SA mechanism.
\end{enumerate}

\begin{figure*} [!t]
    \centering
    \includegraphics[width=1\linewidth]{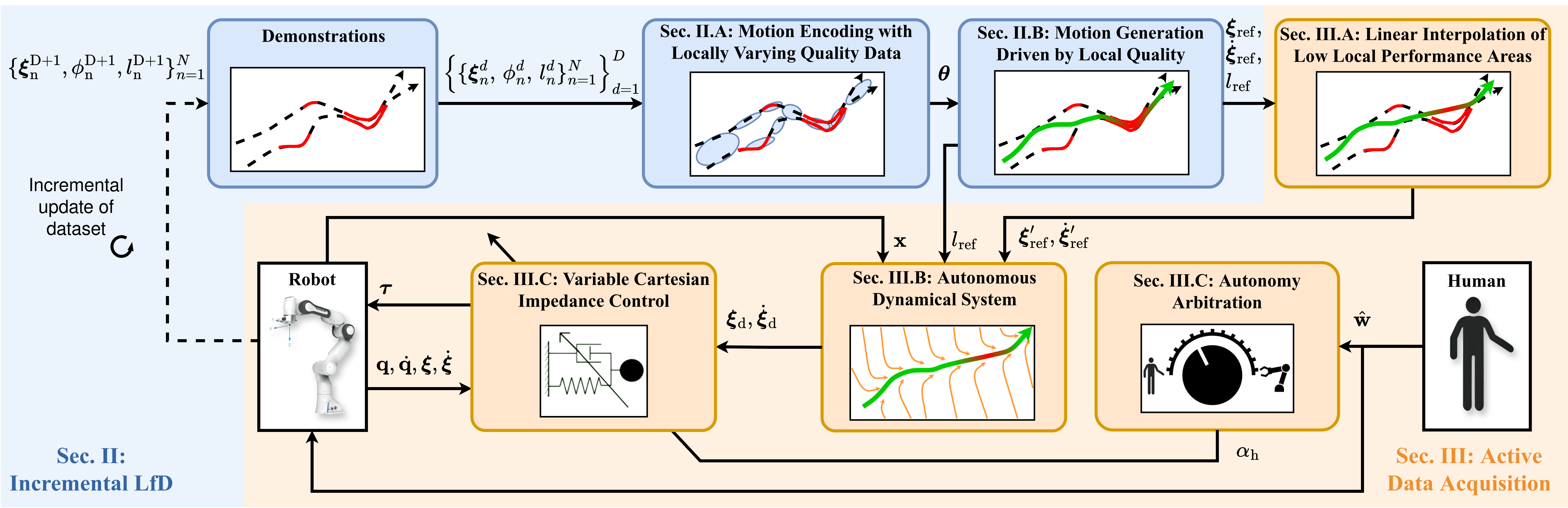}
    \caption{
    Overview of proposed LOPAL framework for active learning from imperfect demonstrations using local demonstration quality information.
    The framework consists of two main components: Incremental LfD from imperfect demonstrations (blue) and active data acquisition (orange).
    The Incremental LfD component is used to encode an incrementally expanded demonstration dataset and generates trajectories guided by local quality information.
    The active data acquisition component enables the robot to collect additional data by executing the generated reference motion while the user provides corrections.
    }
    \vspace{-0.0cm}
    \label{fig:overview}
\end{figure*}

\section{Learning from Locally Imperfect Demonstrations} \label{subsec:perfLfD}
We propose a local performance-aware LfD approach (blue section in Fig. \ref{fig:overview}) to learn efficiently from imperfect demonstrations by leveraging local demonstration quality information.
We encode not only the demonstrated trajectories but also their locally varying quality into a GMM (Sec. II-A).
This multimodal data encoded in the GMM allows generating trajectories conditioned on the local quality information, such that the generated trajectory systematically follows
demonstration samples associated with the highest local performance
(Sec. \ref{subsec:generate}). By combining local high-quality data from multiple demonstrations, the system facilitates the synthesis of behaviors that surpass the quality of any single demonstration. Furthermore, these models can be trained from a single demonstration, enabling immediate robot contribution in subsequent task iterations of the incremental teaching procedure.
See Algorithm 1 for a summary of our local performance-aware LfD approach.

\subsection{Encoding Demonstrations with Locally Varying Quality} \label{subsec:encode}
First, both the demonstrated motion and its local quality information are encoded in a GMM.
Each of the $D$ demonstrations includes $N$ samples of pose $\boldsymbol{\xi} \in \mathbb{R}^6$.
The pose $\boldsymbol{\xi} = \left[ \mathbf{x}, \boldsymbol{\gamma} \right]^T$ is composed of positions $\mathbf{x} \in \mathbb{R}^3$ and orientations in rotation vector representation $\boldsymbol{\gamma} \in \mathbb{R}^3$.
For each sample, a locally varying quality measure is determined, either as a binary penalty $p \in \{0,1\}$ from constraint violations or as a continuous score $p \in [0,1]$ (e.g., reflecting motion smoothness or energy efficiency \cite{zhang_multi-objective_2023}).
These local quality measures are then converted into a smoothed local performance \( l \in [0,1] \) for every sample $n \in \{ 1, \dots , N \}$ using a moving average
\begin{align} \label{eq:locPerf}
    l[n] &= 1 - \frac{1}{\min(N,n+q)-\max(1,n-q)+1}\sum_{i=\max(1,n-q)}^{\min(N,n+q)} p[i].
\end{align}
The half-window size \(q \in \mathbb{N}^+\) balances responsiveness and smoothness when selectively following the locally best demonstration during reference trajectory generation.
To represent the progression of the motion, the phase $\phi  \in [0,1]$
is defined based on the normalized cumulative pose distance.
Finally, the distribution of the data 
$\Bigl\{\{\boldsymbol{\xi}_n^d,\,\phi_n^d,\, l_n^d\}_{n=1}^N\Bigr\}_{d=1}^{D}$
is approximated by training the GMM in an iterative Expectation-Maximization (EM) procedure.
During EM, the log-likelihood of the data is maximized by iteratively updating the GMM parameters
\begin{align*}
    \log \mathcal{L}(\boldsymbol{\theta})
    = \sum_{d=1}^{D}\sum_{n=1}^{N} \log\!\left( \sum_{k=1}^{K} \pi_k \;
    \mathcal{N}\!\bigl(\boldsymbol{\xi}_n^d,\phi_n^d,l_n^d \mid \boldsymbol{\mu}_k,\boldsymbol{\Sigma}_k\bigr)\right),
\end{align*}
where $\mathcal{N}$ denotes the normal distribution.
The GMM with $K$ number of Gaussian components is described by the parameters $\boldsymbol{\theta} = \{ K, \{ \pi_k, \boldsymbol{\mu}_k, \boldsymbol{\Sigma}_k \}_{k=1}^{K}\}$.
Here $\pi_k$ are the priors, $\boldsymbol{\mu}_k$ the means, and $\boldsymbol{\Sigma}_k$ the covariances of the $k$-th component.

\subsection{Generating Trajectories Driven by Local Performance Information} \label{subsec:generate}

Using Gaussian Mixture Regression (GMR) \cite{goswami_learning_2019} we can compute the mean $\bar{\boldsymbol{\mu}}^{\mathbf{b}}(\mathbf{a})$ and variance $\bar{\boldsymbol{\Sigma}}^{\mathbf{b}}(\mathbf{a})$ of the encoded output modalities $\mathbf{b}$ given the input $\mathbf{a}$:
\begin{align*}
    \bar{\boldsymbol{\mu}}^{\mathbf{b}}(\mathbf{a}) &= \sum_{k=1}^{K} h_k(\mathbf{a}) \underbrace{\left( \boldsymbol{\mu}_k^{\mathbf{b}} + \boldsymbol{\Sigma}_k^{\mathbf{b}\mathbf{a}} (\boldsymbol{\Sigma}_k^{\mathbf{a}})^{-1} (\mathbf{a} - \boldsymbol{\mu}_k^{\mathbf{a}}) \right)}_{\hat{\boldsymbol{\mu}}_k^{\mathbf{b}}(\mathbf{a})}, \\
    \bar{\boldsymbol{\Sigma}}^{\mathbf{b}}(\mathbf{a}) &= \sum_{k=1}^{K} h_k(\mathbf{a}) \biggl( \boldsymbol{\Sigma}_k^{\mathbf{b}} - \boldsymbol{\Sigma}_k^{\mathbf{b}\mathbf{a}} (\boldsymbol{\Sigma}_k^{\mathbf{a}})^{-1} \boldsymbol{\Sigma}_k^{\mathbf{a}\mathbf{b}} \\
    & + \left( \hat{\boldsymbol{\mu}}_k^{\mathbf{b}}(\mathbf{a}) - \bar{\boldsymbol{\mu}}^{\mathbf{b}}(\mathbf{a}) \right) \left( \hat{\boldsymbol{\mu}}_k^{\mathbf{b}}(\mathbf{a}) - \bar{\boldsymbol{\mu}}^{\mathbf{b}}(\mathbf{a}) \right) ^\top \biggr).
\end{align*}
Here, $\boldsymbol{\mu}_k^{\mathbf{a}}$ and $\boldsymbol{\mu}_k^{\mathbf{b}}$ are the means of the input and output of the $k$-th Gaussian component, $\boldsymbol{\Sigma}_k^{\mathbf{a}}$ and $\boldsymbol{\Sigma}_k^{\mathbf{b}}$ are the corresponding covariances, and $\boldsymbol{\Sigma}_k^{\mathbf{ba}}$ (and its transpose $\boldsymbol{\Sigma}_k^{\mathbf{ab}}$) denotes the cross-covariance between output and input modalities.
The normalized responsibilities $h_k(\mathbf{a})$ are computed as
\begin{align*}
    h_k(\mathbf{a}) = \frac{\pi_k \, \mathcal{N}(\mathbf{a} \mid \boldsymbol{\mu}_k^{\mathbf{a}}, \boldsymbol{\Sigma}_k^{\mathbf{a}})}{\sum_{i=1}^{K} \pi_i \, \mathcal{N}(\mathbf{a} \mid \boldsymbol{\mu}_i^{\mathbf{a}}, \boldsymbol{\Sigma}_i^{\mathbf{a}})}.
\end{align*}
Thus, for each of the $M$ reference phase values $\phi_{{\rm ref},m}$, $m \in \{1, \dots, M\}$, defined as a sequence linearly increasing from 0 to 1, a phase-dependent performance target can be extracted as
\begin{align} \label{eq:ref_perf}
    l_{{\rm ref},m} = \bar{\mu}^{l}(\phi_{{\rm ref},m}) + 2 \sqrt{\bar{\Sigma}^{l}(\phi_{{\rm ref},m})}. %
\end{align}
This sets the performance target at two standard deviations above the phase-dependent mean to ensure a high reference performance level; to obtain more conservative performance targets, the mean-offset could be reduced.
Using the reference phase and this performance target, we regress the reference pose $\boldsymbol{\xi}_{{\rm ref},m} \in \mathbb{R}^6$ as
\begin{align} \label{eq:gmr}
    \boldsymbol{\xi}_{{\rm ref},m} = \bar{\boldsymbol{\mu}}^{\boldsymbol{\xi}}([\phi_{{\rm ref},m},\, l_{{\rm ref},m}]^\top)
\end{align}
By conditioning on the high performance target, the generated movement selectively follows the locally highest-performing demonstrations, thereby preventing degradation through suboptimal demonstrations.
Defining the phase via normalized cumulative pose distance and setting $\phi_{\rm ref}$ to increase linearly ensures that the generated motion maintains nonzero velocity, even when the demonstrations contain pauses.

\begin{algorithm}[b] \label{alg:lfd}
    \caption{\small Local Performance-Driven LfD}
    \begin{algorithmic}[1]
        \STATE \textbf{Input:} Demonstration dataset $\{\{\boldsymbol{\xi}_n^d = [\mathbf{x}_n^d, \boldsymbol{\gamma}_n^d], \phi_n^d, p_n^d\}_{n=1}^N\}_{d=1}^D$
        \STATE \hspace{1.65em} Parameters: $K$ (GMM components), $q$ (half-window size)
        \STATE \textbf{Output:} Reference trajectory $\{\boldsymbol{\xi}_{\text{ref},m}, l_{\text{ref},m}\}_{m=1}^M$
        \STATE \textbf{===== ENCODING DEMONSTRATIONS =====}
        \FOR{each demonstration $d \in \{1,\dots,D\}$}
            \STATE $\phi^d \gets$ ComputePhase($\{\boldsymbol{\xi}_n^d\}$)
            \STATE $l^d \gets$ ComputeLocalPerformance($\{p_n^d\}$, $q$) \qquad \qquad \qquad // eq.~\eqref{eq:locPerf}
        \ENDFOR
        \STATE $\{\boldsymbol{\xi}_n^d, \phi_n^d, l_n^d\} \gets$ AlignDemonstrations($\{\boldsymbol{\xi}_n^d\}$)
        \STATE $\{\boldsymbol{\xi}_n^d, \phi_n^d, l_n^d\} \gets$ NormalizeData($\{\boldsymbol{\xi}_n^d, \phi_n^d, l_n^d\}$)
        \STATE $\theta \gets$ TrainGMM($\{\boldsymbol{\xi}_n^d, \phi_n^d, l_n^d\}$, $K$)     
        \STATE \textbf{===== GENERATING REFERENCE TRAJECTORIES =====}
        \FOR{each reference phase $\phi_{\text{ref},m} \in \text{linspace}(0,1,M)$}
            \STATE $l_{\text{ref},m} \gets$ ExtractTargetPerformance$(\phi_{\text{ref},m}, \theta)$ \qquad \qquad \; \; // eq.~\eqref{eq:ref_perf}
            \STATE $\boldsymbol{\xi}_{\text{ref},m} \gets$ GMR($\theta$, $\phi_{\text{ref},m}$, $l_{\text{ref},m}$)  \qquad \qquad \qquad \qquad \qquad \, // eq.~\eqref{eq:gmr}
        \ENDFOR  
        \STATE \textbf{return} $\{\boldsymbol{\xi}_{\text{ref},m}, l_{\text{ref},m}\}_{m=1}^M$
    \end{algorithmic}
\end{algorithm}
\section{Active Data Acquisition Guided by Local Performance Information} \label{subsec:activeData}
The active data acquisition component (orange section in Fig. \ref{fig:overview}) incrementally expands the demonstration dataset, enabling the LfD approach to learn behaviors of higher performance.
To request better demonstrations in areas where good data is missing, the robot reduces the speed of the motion execution (Sec. \ref{subsec:autoDS}) and gives a visual indication (e.g. LED light). The user can dynamically take control to make the necessary corrections by applying wrench to the robot, which results in the robot increasing its control compliance (Sec. \ref{subsec:arbitrate}). To potentially reduce the required corrections, the reference trajectory is modulated in low-quality areas through interpolation, which leverages neighboring high-quality data (Sec. \ref{subsec:interpolation}).
In well-demonstrated areas, the robot can accurately reproduce the learned motion of high local performance. Thus, corrective human input can be focused precisely where the robot has difficulties. 
See Algorithm 2 for a summary of the active data acquisition approach.

\subsection{Performance-Guided Local Motion Modulation}
\label{subsec:interpolation}
In low-performance regions, we leverage neighboring high-quality information to create better nominal behavior and reduce the need for user corrections (Fig. \ref{fig:hrisetup}).
If the reference local performance $l_{\text{ref},m}$ of sample $m \in \{1, \dots, M\}$
is below a threshold value $l_{\text{th}} \in [0,1]$, its position is replaced through linear interpolation between the closest preceding and succeeding high-quality samples:
\begin{align} \label{eq:interp}
\mathbf{x}'_{\text{ref},m} = \mathbf{x}_{\text{ref},m^-} + \frac{\phi_{\text{ref},m}-\phi_{\text{ref},m^-}}{\phi_{\text{ref},m^+}-\phi_{\text{ref},m^-}}\,\bigl(\mathbf{x}_{\text{ref},m^+}-\mathbf{x}_{\text{ref},m^-}\bigr)
\end{align}
where $m^- < m$ and $m^+ >m$ denote the indices of the closest samples with $l_{\text{ref},m^\pm} \ge l_{\text{th}}$. 
This interpolation ensures motions in low local quality areas are modulated using neighboring high-quality samples, while preserving the original behavior in well-demonstrated areas.
To ensure a smooth trajectory, a second-order Butterworth low-pass filter with a cutoff frequency of \SI{2}{\Hz} is applied. We compute the reference twist $\dot{\boldsymbol{\xi}}'_{\text{ref},m} = \left[ \dot{\mathbf{x}}_{{\rm ref}, m}', \boldsymbol{\omega}'_{{\rm ref}, m} \right]^T \in \mathbb{R}^6$ by numerically differentiating the regressed reference pose trajectory, where $\boldsymbol{\omega}'_{{\rm ref}, m} \in \mathbb{R}^3$ is the reference angular velocity.

\subsection{Executing Motions with an Autonomous Dynamical System} \label{subsec:autoDS}
To accurately reproduce the reference motion with the robot in a state-dependent manner, an autonomous dynamical system is used. A proportional feedback control term facilitates attraction towards the reference pose trajectory, while a feedforward reference twist drives the system forward

\begin{align} \label{eq:motion_control}
    \dot{\boldsymbol{\xi}}_{\rm d} &= \mathbf{K}_{DS} \begin{bmatrix} \mathbf{x}_{{\rm ref},j}' - \mathbf{x} \\ \boldsymbol{e}_{\rm rot}(\boldsymbol{\gamma}_{{\rm ref},j}', \boldsymbol{\gamma}) \end{bmatrix} + s \dot{\boldsymbol{\xi}}'_{{\rm ref},j},
\end{align}
where $\boldsymbol{e}_{\rm rot}(\boldsymbol{\gamma}_1, \boldsymbol{\gamma}_2) \in \mathbb{R}^3$ is the minimal rotation from $\boldsymbol{\gamma}_2$ to $\boldsymbol{\gamma}_1$ in rotation vector representation.
The streamlines in Fig. \ref{fig:overview} illustrate the resulting potential field of the autonomous dynamical system.
The attraction stiffness is defined as $\mathbf{K}_{DS} = \operatorname{diag}(k_{DS, {\rm t}} \;\mathbf{I}_3, \; k_{DS, {\rm r}}\,\mathbf{I}_3) \in \mathbb{R}^{6 \times 6}$, where $k_{DS,{\rm t}} \in \mathbb{R}^+$ and $k_{DS,{\rm r}} \in \mathbb{R}^+$ are the translational and rotational stiffness gains.
The reference twist in eq. (\ref{eq:motion_control}) is scaled by the factor $s \in \left[s_{\rm min}, 1\right]$ to slow down the motion progression when entering regions of low local performance:
\begin{align} \label{eq:slowdown}
    s &= s_{\min} + (1 - s_{\min})  \left( \min_{j'  \in \{j,\dots, \min(j+\delta, M)\}} (l_{{\rm ref},j'}) \right).
\end{align} 
Here, $\delta \in \mathbb{N}^+$ is the lookahead window size allowing to proactively slow down before entering the low-performance region, and $s_{\rm min} \in [0,1]$ is the minimum velocity scale-down factor.
The reduced velocity indicates to the user that the robot is struggling to execute this part of the motion well and gives the human more reaction time to correct the robot's motion. 
Further, the $s$ value can be used to give additional feedback about low-performance areas to the user, e.g. in the form of visual cues.
\\
The index \(j \in \{1, \dots,M\}\) identifies the reference pose $\boldsymbol{\xi}_{{\rm ref}}' \in \mathbb{R}^6$ closest to the current pose \(\boldsymbol{\xi} \in \mathbb{R}^6\) to couple motion progression with the pose alignment. This ensures local attraction to the reference motion and makes the dynamical system time-invariant. 
The k-d tree algorithm is used to efficiently compute this pose distance.

\begin{algorithm}[b]
    \caption{\small Performance-Guided Active Data Acquisition}
    \label{alg:ada}
    \begin{algorithmic}[1]
        \STATE \textbf{Input:} Reference trajectory $\{\boldsymbol{\xi}_{\text{ref},m}, l_{\text{ref},m}\}_{m=1}^M$, wrench $\hat{\mathbf{w}}$
        \STATE \hspace{1.65em} Parameters: $l_{\text{th}}$ (interpolation threshold), $\mathbf{K}_{DS}$ (stiffness of dynamical system), $s_{\min}$ (min speed), $\delta$ (lookahead), $a,b,c$ (autonomy activation parameters), $r^+, r^-, r_q^-$ (autonomy filter parameters), $\mathbf{K}_{\rm imp}$ (impedance controller stiffness)
        \STATE \textbf{Output:} New demonstration $\{\boldsymbol{\xi}_n^{\text{new}}, p_n^{\text{new}}\}_{n=1}^{N^{\text{new}}}$
        \STATE \textbf{===== INTERPOLATE \& FILTER REFERENCE =====}
        \STATE $\{\mathbf{x}'_{\text{ref},m}\} \gets$ InterpolateLowPerf(\{$\mathbf{x}_{\text{ref},m}$, $\phi_{\text{ref},m}$, $l_{\text{ref},m}\}$, $l_{\text{th}}$) \hfill // eq.~\eqref{eq:interp}
        \STATE $\{\boldsymbol{\xi}'_{\text{ref},m}\} \gets$ LowPassButterworth($\{[\mathbf{x}'_{\text{ref},m},\, \boldsymbol{\gamma}_{\text{ref},m}]^T\}$, cutoff $= \SI{2}{\Hz}$)
        \STATE $\{\dot{\boldsymbol{\xi}}'_{\text{ref},m}\} \gets$ NumericallyDifferentiate($\{\boldsymbol{\xi}'_{\text{ref},m}\}$)
        \STATE Initialize: $j \gets 1$, $\alpha_h[0] \gets 0$, $\boldsymbol{\xi}_{\text{new}} \gets \{\}$, $p_{\text{new}} \gets \{\}$
        \WHILE{$j < M$}
            \STATE \textbf{===== AUTONOMOUS MOTION TRACKING =====}
            \STATE $j \gets$ FindClosestReference($\boldsymbol{\xi}$, $\{\boldsymbol{\xi}'_{\text{ref},m}\}$)
            \STATE $s \gets$ ComputeSpeedScale($\{l_{\text{ref}}\}$, $j$, $\delta$, $s_{\min}$) \hfill // eq.~\eqref{eq:slowdown}
            \STATE $\dot{\boldsymbol{\xi}}_{\rm d} \gets \mathbf{K}_{DS} (\boldsymbol{\xi}'_{\text{ref},j} - \boldsymbol{\xi}) + s \dot{\boldsymbol{\xi}}'_{\text{ref},j}$ \hfill // eq.~\eqref{eq:motion_control}
            \STATE \textbf{===== AUTONOMY ARBITRATION =====}
            \STATE $\alpha_{\mathbf{w}} \gets$ ComputeWrenchAutonomy($\hat{\mathbf{w}}$, $a$, $b$, $c$) \hfill // eq.~\eqref{eq:alpha_force}
            \STATE $\alpha_h \gets$ FilterAutonomy($\alpha_h$, $\alpha_{\mathbf{w}}$, $r^+$, $r^-$, $r_q^-$) \hfill // eq.~\eqref{eq:alpha_filt}
            \STATE $\mathbf{K} \gets (1 - \alpha_h)\mathbf{K}_{\text{imp}}$, $\mathbf{D} \gets 2\sqrt{\mathbf{K}}$ \hfill // Adjust control compliance
            \STATE \textbf{===== DATA ACQUISITION =====}
            \STATE ExecuteControl($\dot{\boldsymbol{\xi}}_{\rm d}$, $\mathbf{K}$, $\mathbf{D}$)
            \STATE $p \gets$ EvaluatePenalty
            \STATE $\boldsymbol{\xi}_{\text{new}} \gets \boldsymbol{\xi}_{\text{new}} \cup \{\boldsymbol{\xi}\}$ \qquad \qquad \qquad \qquad \qquad \qquad \quad  // Record pose
            \STATE $p_{\text{new}} \gets p_{\text{new}} \cup \{p\}$ \hfill // Record penalty
        \ENDWHILE
        \STATE \textbf{return} $\{\boldsymbol{\xi}_n^{\text{new}}, p_n^{\text{new}}\}_{n=1}^{N^{\text{new}}}$ \hfill // Expand dataset
    \end{algorithmic}
\end{algorithm}

\begin{figure*} [t]
    \centering
    \includegraphics[width=0.95\linewidth]{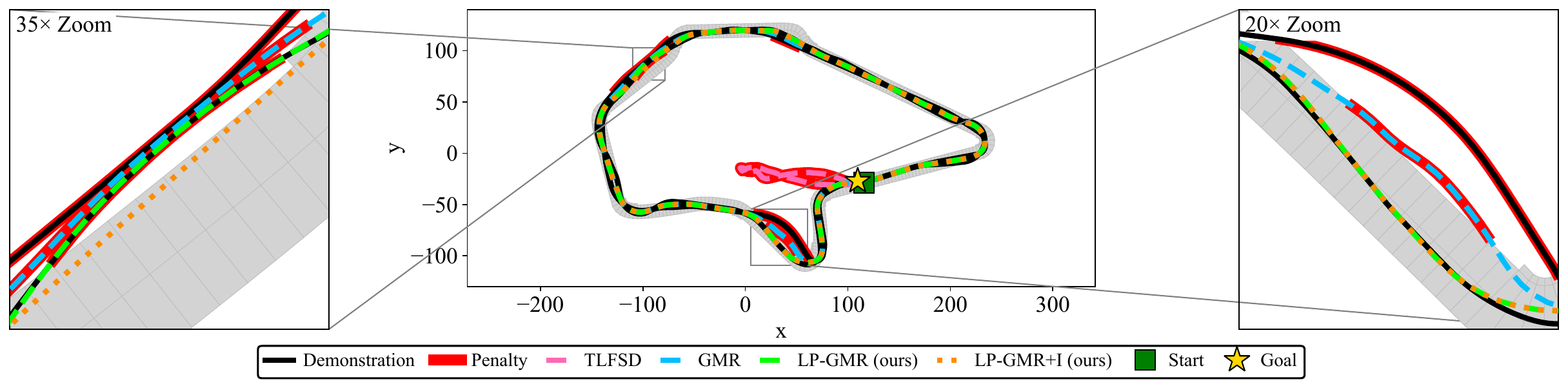}
    \vspace{-0.0cm}
    \caption{Illustration of the car driving simulation with two exemplary demonstrations (black). The penalties (red) occurred when the car left the track. The averaging behavior of the GMR (blue dashed), which purely encodes the demonstrated motion, could cause the generated movement to result in penalties, even though one of the demonstrations had shown how to avoid these penalties. 
    Due to these local failures, the motion generated with TLFSD (pink) maximizes distance to the imperfect demonstrations, which often results in even more penalties than in the demonstrated motions.
    In the proposed local performance-aware approach (green dashed), the information about the location of penalties was effectively used to avoid these penalty areas. 
    Experiment results demonstrated that interpolating regions lacking high quality demonstration data (orange dotted) allowed to further reduce penalties.
    }
    \label{fig:car_racing_track}
    \vspace{-0.0cm}
\end{figure*}

\subsection{User Corrections through Autonomy Arbitration} \label{subsec:arbitrate}
A shared autonomy approach can allow the human to selectively take over autonomy during incremental teaching in order to correct the robot's task execution.
To facilitate human intervention in our LOPAL framework, the robot dynamically adjusts its control compliance to respond to human-applied forces and torques (see Fig. \ref{fig:overview}).
This autonomy arbitration through compliance modulation is controlled by the human autonomy $\alpha_{\rm h} \in (0,1)$, which is calculated by filtering the current wrench-based autonomy value $\alpha_{\mathbf{w}} \in (0,1)$ to enable smooth and adaptable autonomy arbitration.
First, we compute the value $\alpha_{\mathbf{w}}$ using the wrench $\hat{\mathbf{w}} \in \mathbb{R}^6$ \cite{jadav_shared_2024}:
\begin{align} \label{eq:alpha_force}
    \alpha_{\mathbf{w}} = 0.5\Biggl(1+\tanh\left(\frac{3}{c \, a}(\|\hat{\mathbf{w}}\| - \frac{a}{2} - b)\right)\Biggr).
\end{align}
Here, $a$ controls how quickly the activation increases with $\|\hat{\mathbf{w}}\|$, and $b$ sets the deadzone below which the response is suppressed. To enforce the tolerance $\alpha_{\mathbf w}(\|\hat{\mathbf w}\|=b)=1-\alpha_{\mathbf w}(\|\hat{\mathbf w}\|=a+b)=\epsilon$, we choose $c=\dfrac{3}{2\,\operatorname{atanh}(1-2\epsilon)}$.
The wrench $\hat{\mathbf{w}}$ is obtained from a force–torque sensor mounted at the robot wrist, which enables high sensitivity to human-applied interaction wrench.

To avoid discontinuities, a recursive filter is used to compute $\alpha_{h}$ at the $i$-th control loop step:
\begin{align} \label{eq:alpha_filt}
    \alpha_{h}[i] &= \alpha_{h}[i-1] + r_{\alpha}[i]\bigl(\alpha_{\mathbf{w}}[i] - \alpha_{h}[i-1]\bigr).
\end{align}
where the filter gain \(r_{\alpha} \in \mathbb{R}^{+}\) is defined as
\begin{align} \label{eq:alpha_filt_gain}
    r_{\alpha}[i] = 
    \begin{cases}
        r^{+}, & \text{if } {\alpha}_{\mathbf{w}}[i] > {\alpha}_{{\mathbf{w}}}[i-1], \\ r^{-} + r^{-}_{q} ( 1 - {\alpha}_{{\mathbf{w}}}[i] )^2, & \text{otherwise}.
    \end{cases}
\end{align}
To achieve a rapid human autonomy increase when force is applied and a slower decrease when force is released, we set $r^{+} > r^{-}$.
The quadratic term in eq. (\ref{eq:alpha_filt_gain}) with gain $r_q^-$ facilitates a fast reduction for small $\alpha_{{\mathbf{w}}}$ values, which can result from briefly applied low forces that may be transient and less indicative of a persistent human intention to take control.
Finally, $\alpha_{h}$ is used to modulate the stiffness $\mathbf{K} = (1 - \alpha_{\rm h})\mathbf{K}_{\rm imp} = (1 - \alpha_{\rm h}) \operatorname{diag}(k_{{\rm t}} \;\mathbf{I}_3, \; k_{{\rm r}}\,\mathbf{I}_3) \in \mathbb{R}^{6 \times 6}$ and damping $\mathbf{D} = 2\sqrt{\mathbf{K}} \in \mathbb{R}^{6 \times 6}$ of a Cartesian variable impedance controller \cite{xue_shared_2023, jadav_shared_2024}. 
An energy-tank-based approach \cite{jadav_shared_2024} was used to ensure the passivity of the controller and thus increase safety when operating in close proximity to humans \cite{michel_novel_2024}.

\section{Simulation Evaluation} \label{subsec:sim}
We evaluated our local performance-aware LfD method on a car driving task using human demonstrations on a 2D track. Specifically, we investigated the ability to generate trajectories that avoid local penalties while learning efficiently from an incrementally expanding set of demonstrations. In this work, all statistical analyses were performed using Wilcoxon signed-rank tests, as the data were paired and not normally distributed.
Effect sizes are reported as rank-biserial correlation (r), where 0 indicates no difference and $\pm 1$ indicates one condition consistently ranked higher than the other, with 95\% bootstrap confidence intervals (CI).
\subsection{Simulation Setup}
The task required driving on a 2D track while avoiding penalties, which were incurred whenever the vehicle's center left the road.
The demonstrations were collected by controlling the car via keyboard on a fixed track in the OpenAI Gym car racing environment.
The training data comprised 15 demonstrations, each containing a 2D position trajectory with corresponding phase and local performance values.
In 50 experimental sessions, each time we started with one of the demonstrations and gradually added more demonstrations one-by-one to the training dataset in random order. This allowed us to assess the ability to learn efficiently from an incrementally expanded set of imperfect demonstrations.

\subsection{Compared LfD Methods} \label{subsec:comparedLfDMethods}
\subsubsection{Reinforcement Learning from Prior Data (RLPD) \cite{ball_efficient_2023}}
Learning combines LfD and RL by sampling equally from a demonstration buffer and an online replay buffer during training. Local performance information is used as a dense reward.
\subsubsection{Trajectory Learning from Failed and Successful Demonstrations (TLFSD) \cite{hertel_learning_2021}}
Demonstration quality is assessed globally by classifying each demonstration as successful or unsuccessful. Separate GMMs are trained for each set, and trajectories are generated by minimizing deviation from the successful GMM while maximizing deviation from the failed GMM using Mahalanobis distance.
\subsubsection{Vanilla GMM/GMR (GMR)}
The GMM is trained on motion data, while the performance information is excluded from the training data. Consequently, motion generation is performed using only the phase variable as regression input.
\subsubsection{Local Performance Pruned GMM/GMR (P-GMR)}
Local performance information is used to prune the penalized samples from the training data such that the GMM is trained only on high-quality motion data.
Motion generation uses the phase variable as regression input, as in the vanilla GMR.
\subsubsection{Local Performance-Aware GMM/GMR (LP-GMR)}
Local performance information is encoded in the GMM together with the motion data. 
This allows to condition the motion generation both on phase and on high local performance such that the reference motion follows the locally best available motion (Sec.~\ref{subsec:generate}).
\subsubsection{Local Performance-Aware GMM/GMR with Interpolation (LP-GMR+I)}
Extending LP-GMR, the encoded local performance information is used not only for motion regression but also to interpolate motions in regions lacking high-quality demonstration data (Sec.~\ref{subsec:interpolation}).

\begin{figure} [t]
    \vspace{-0.0cm}
    \centering
    \includegraphics[width=0.99\linewidth]{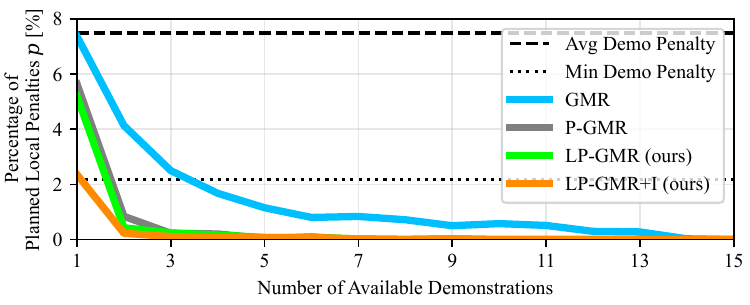}
    \vspace{-0.4cm}
    \caption{Reduction of penalties for generated motions as an increasing number of demonstrations become available (fewer penalties are better). 
    Both ablations using local performance information (LP-GMR green and P-GMR gray) achieved better performance using fewer demonstrations compared to the base GMR approach (blue) which was not informed by demonstration quality. Additional performance benefits were achieved by interpolating regions lacking high-quality demonstration data (LP-GMR+I orange).
    As a reference, the average (black dashed) and minimum (black dotted) penalties from demonstrations are shown.
    }
    \vspace{-0.0cm}
    \label{fig:car_racing_results}
\end{figure}
\vspace{-0.0cm}
\subsection{Simulation Results}
If at least one successful (i.e. penalty-free) demonstration exists, TLFSD can replicate the motion without penalties. 
However, for challenging tasks, especially with long horizons, it can be difficult to provide a full demonstration without a single penalty, which would result in the demonstration being labeled as failed.
If only failed demonstrations are available, TLFSD often generates trajectories with many penalties, since it maximizes the distance from these imperfect demonstrations (pink trajectory in Fig. \ref{fig:car_racing_track}).
While iterative generative learning can improve results, it requires careful parameter tuning and does not guarantee a successful motion within a fixed number of iterations, particularly for long or complex motions.
All 15 demonstrations exhibited imperfections on the full track. 
Since TLFSD generated even more penalties than these demonstrations, it was not considered in the detailed evaluation on the full track (Fig. \ref{fig:car_racing_results}).
Unlike TLFSD, which relies on global assessment of demonstration success, our approach leverages local quality cues to handle partial imperfections. 
The LfD approaches leveraging local performance information can generate motions that selectively follow the locally best available demonstrated behavior (see right zoomed view of Fig. \ref{fig:car_racing_track}).
Conversely, base GMR, which does not leverage any performance information, is prone to replicate low-performance behaviors present in most demonstrations due to its averaging nature.
If all demonstrations show penalties in a region, interpolating between neighboring high-quality samples can reduce or even avoid these penalties, as illustrated in the left zoomed view of Fig. \ref{fig:car_racing_track}.
The LfD methods leveraging local performance information \rev{(P-GMR and LP-GMR)} yielded significantly fewer penalties than the base GMR ($p<0.001$, smallest $|r| = 0.98$, CI$_{95\%}$=[0.95, 0.99]). This can be attributed to their capability to utilize complementary high-quality local demonstration data.
\rev{As mentioned in Sec.~\ref{subsec:encode}, local performance can be derived not only from binary penalties, as in this task, but also from continuous ratings. In contrast to LP-GMR, when applying P-GMR to such continuous ratings, there is no principled criterion for choosing the pruning threshold, and the retained samples contribute equally to motion generation regardless of their differing quality levels.}
Compared to the performance leveraging ablations, linear interpolation of low-performance regions (LP-GMR+I) further improved learning efficiency when using only one or two demonstrations, resulting in an additional penalty reduction ($p<0.006$, smallest $|r| = 0.73$, CI$_{95\%}$=[0.38, 0.97]).
RLPD achieved penalty-free task performance but required $430 \pm 140$ (mean $\pm$ std) additional exploration episodes ($129{,}013 \pm 46{,}852$ steps); \rev{such extensive exploration may not pose a problem in simulation but could be impractical or potentially unsafe in real-world deployment.}
\\
These results underscore the capability of the proposed LfD approach to enhance learning efficiency from imperfect demonstrations by using local performance information.

\begin{figure} [t]
    \vspace{-0.0cm}
    \centering
    \includegraphics[width=0.83\linewidth]{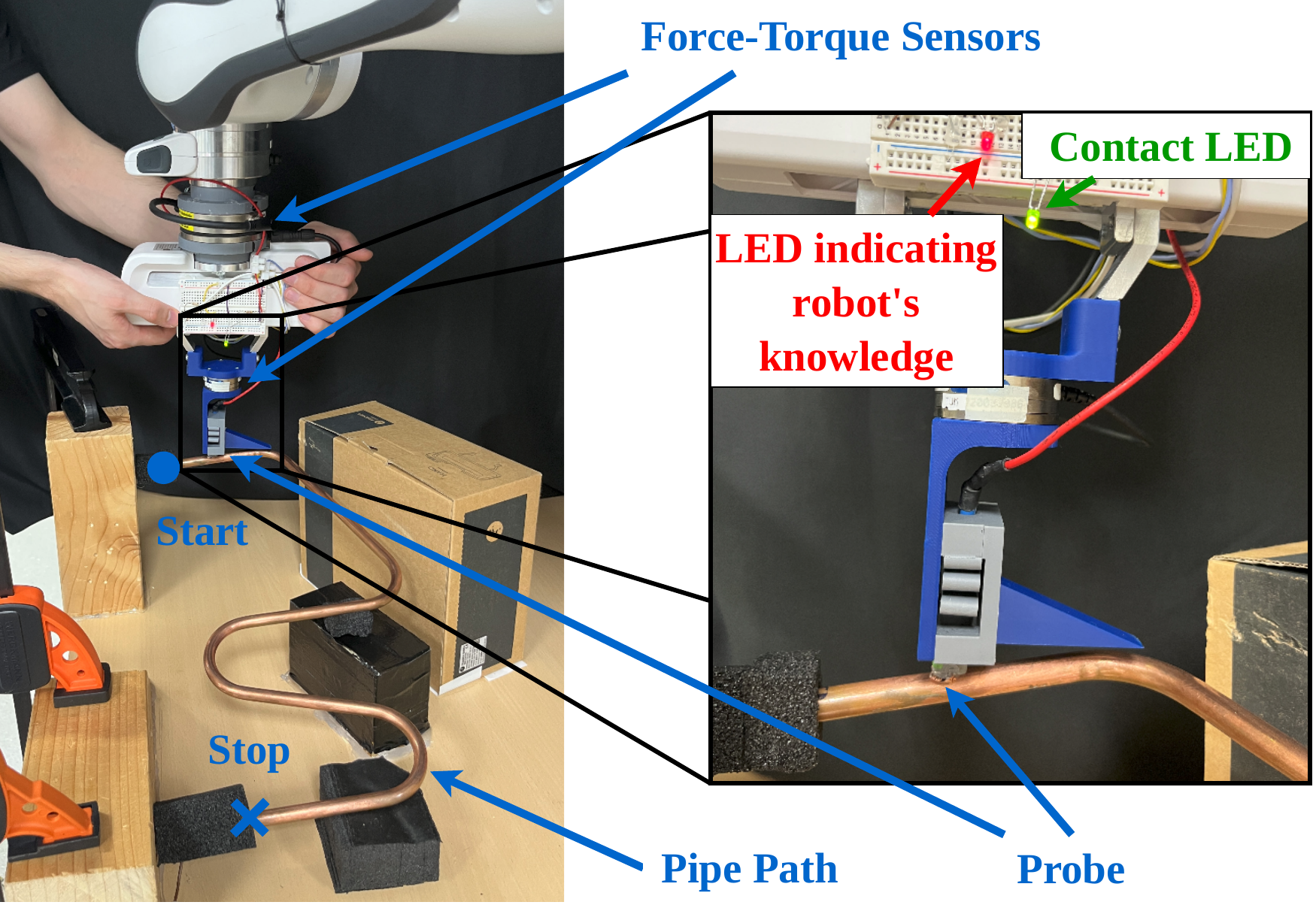}
    \caption{
    Pipe inspection experiment setup used for the validation of LOPAL.
    Here, participants were tasked to trace the pipe path with the probe from start to end, while maintaining contact between probe and pipe which is indicated by the green LED. In one of the teaching modes, a red LED indicates the robot's limited knowledge of high-quality data to the user. Two force-torque sensors are used to determine the forces applied by the user.}
    \vspace{-0.0cm}
    \label{fig:experiment_setup}
\end{figure}

\section{Real-World Experiment}
We evaluated our proposed LOPAL approach in a real-world incremental teaching scenario on a Franka Research 3 robot. 
Specifically, we investigated how teaching with the proposed method affects the burden on the user.
In addition, we evaluated our proposed LfD approach in this real-world experiment, similar to the evaluation performed for the simulation.
The experiment was conducted with 8 participants (no prior power analysis), who confirmed their participation in the experiment by signing a written informed consent form. \rev{The ethics proposal for this study was approved by the TU Wien Research Ethics Committee (case number 045\_19012024\_TUWREC).}

\subsection{Experiment Setup}
In this real-world experiment, users had to kinesthetically teach the robot a pipe inspection task.
Participants were asked to trace a copper pipe while maintaining contact between the end-effector-mounted probe and the pipe, see Fig. \ref{fig:experiment_setup}.
The loss of contact between pipe and probe was considered a penalty.
Maintaining continued contact is important because sensors used for inspection often require direct contact or a fixed offset for proper functioning.
Because the pipe was mounted in an inclined orientation and supporting structures had to be avoided, the position and orientation of the end effector needed to be controlled in order to trace the pipe path without losing contact.
For every teaching mode, an initial demonstration with low performance was provided such that in the shared autonomy approaches the robot could contribute from the first trial and the starting behavior was the same for all participants, making results more comparable. 
In all teaching modes, three demonstrations were collected sequentially, incrementally expanding the dataset and updating the model after each trial.
After collecting the demonstrations with each teaching mode, participants assessed their perceived mental and physical workload using the NASA-TLX questionnaire.
To compare different LfD methods (Sec. \ref{subsec:comparedLfDMethods}) regarding their efficiency to learn from imperfect demonstrations, the robot autonomously executed the generated behaviors without user corrections while the demonstration set was incrementally expanded (evaluated on six datasets of four demonstrations).
Performance in this task was evaluated in terms of the percentage of pipe path traced without penalty.
The forces exerted by the user were determined by comparing the sensor values of the two force-torque sensors shown in Fig. \ref{fig:experiment_setup}.
The parameters used for our LOPAL approach in this experiment are summarized in Table \ref{tab:exp_params}.
\begin{figure}[t]
    \vspace{-0.0cm}
    \centering
    \includegraphics[width=0.97\linewidth]{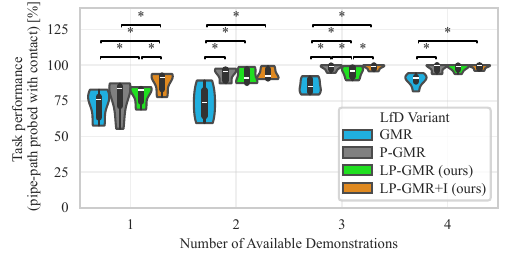}
    \vspace{-0.0cm}
    \caption{Comparison of LfD methods in terms of task performance as the number of demonstrations increased incrementally. %
    Leveraging local performance information (P-GMR gray, LP-GMR green, LP-GMR+I orange) allowed to achieve higher task performance with fewer demonstrations compared to the GMR baseline (blue). When only a single demonstration was available, interpolation of low-quality regions (LP-GMR+I) provided further performance benefits.}
    \label{fig:experiment_lfd}
    \vspace{-0.0cm}
\end{figure}

\begin{table}[!t]
    \vspace{0.0cm}
    \centering
    \caption{The empirically selected parameters used for the proposed LOPAL framework in the real-world experiment.}
    \label{tab:exp_params}
    \small
    \begin{tabular}{|l|l|l|l|}
    \hline
    $r^+=0.03$ & $k_{DS,\text{t}}=\SI{47}{\per\second}$ & $k_{\text{t}}=\SI{2500}{\newton\per\meter}$ & $a=10$ \\
    \hline
    $r^-=0.002$ & $k_{DS,\text{r}}=\SI{27}{\per\second}$ & $k_{\text{r}}=\SI{15}{\newton\meter\per\radian}$ & $b=3$ \\
    \hline
    $r^-_q=0.02$ & $K=95$ & $f_c = \SI{150}{\Hz}$ & $c=0.434$ \\
    \hline
    $s_{\rm min} = 0.4$ & $\delta = 45$ & $q=3$ & $l_{\rm th} = 0.7$ \\
    \hline
    \end{tabular}
    \vspace{-0.0cm}
\end{table}
\subsection{Compared kinesthetic teaching methods}
\subsubsection{Gravity Compensation (GC)}
The robot remains passive such that the human is in full control throughout the task execution.
\subsubsection{Shared Autonomy with GMR (SA + GMR)}
The robot executes the behavior which was learned without leveraging performance information (GMR). By applying wrench to the robot, the user can dynamically take control to correct the robot's task execution (Sec. \ref{subsec:arbitrate}).
\subsubsection{Shared Autonomy with P-GMR (SA + P-GMR)}
The robot executes the motion learned from pruned, high-quality data (P-GMR).
Our proposed SA method allows the user to correct the robot's task execution.
\subsubsection{LOPAL (ours)}
Our wrench-based shared autonomy method is combined with the proposed LfD method (LP-GMR+I). 
During teaching, active performance feedback is provided through speed modulation and visual indication (red LED, Fig.~\ref{fig:experiment_setup}) to prompt user corrections.

\subsection{Experiment Results}
\subsubsection{Comparison of LfD approaches}
Figure \ref{fig:experiment_lfd} illustrates how task performance evolves as the number of available demonstrations increases.
In this real-world experiment, we compared the LfD methods GMR, P-GMR, LP-GMR, and our LP-GMR+I (described in Sec.~\ref{subsec:comparedLfDMethods}).

Consistent with the simulation results, the real-world experiment results also showed that all LfD approaches leveraging local performance information tended to achieve better performance than baseline GMR. Our proposed approach (LP-GMR+I) consistently achieved better performance than base GMR ($p=0.031$, $|r| = 1$, CI$_{95\%}$=[$1.00$, $1.00$]), achieving an average performance improvement of up to $27.31\%$ depending on the number of demonstrations.
When only a single demonstration was available, LP-GMR+I's interpolation mechanism reduced the low-performance areas further compared to P-GMR and LP-GMR ($p=0.031$, smallest $|r| = 1$, CI$_{95\%}$=[$1.00$, $1.00$]).

\begin{figure}[!t]
    \vspace{-0.0cm}
    \centering
    \includegraphics[width=0.999\linewidth]{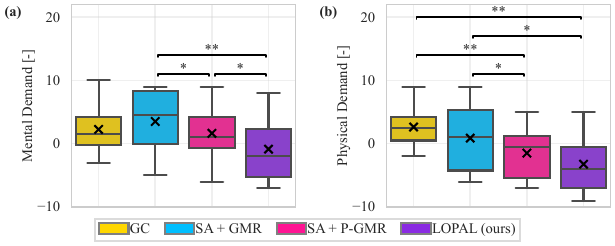}
    \vspace{-0.4cm}
    \caption{Subjective ratings of \textbf{(a)} mental and \textbf{(b)} physical demand for the different teaching variants: GC (yellow), SA + GMR (blue), SA + P-GMR (pink), and LOPAL (purple).}
    \label{fig:questionnaire}
    \vspace{-0.0cm}
\end{figure}

\subsubsection{Comparison of Teaching Approaches}
\begin{figure}[b]
    \vspace{-0.0cm}
    \centering
    \includegraphics[width=0.99\linewidth]{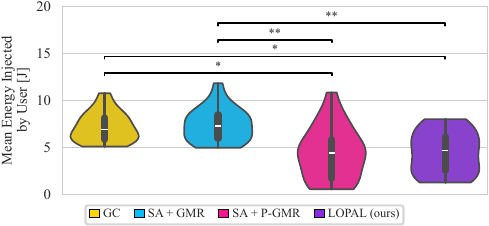}
    \vspace{-0.4cm}
    \caption{Mean energy injected by the participants during the teaching trials when using different teaching variants. When incrementally teaching with approaches that leverage local performance information (SA + P-GMR pink and LOPAL purple), users expended less energy than with GC (yellow) and SA + GMR (blue), which do not use local performance information.}
    \label{fig:mean_user_energy}
    \vspace{0.0cm}
\end{figure}

As illustrated in Fig. \ref{fig:questionnaire} \textbf{(b)}, participants reported reduced physical task load with the teaching approaches that leverage local performance information (SA + P-GMR and LOPAL) compared to baselines (GC and SA + GMR) which do not use performance information ($p<0.028$, smallest $|r| = 0.83$, CI$_{95\%}$=[$0.21$, $0.97$]).
These subjective results are also reflected quantitatively, as teaching with GC and SA + GMR resulted in higher energy injected by the user ($p<0.023$, smallest $|r| = 0.89$, CI$_{95\%}$=[$0.50$, $1.00$]), see Fig. \ref{fig:mean_user_energy}.
The high physical demand with these two approaches may be due to the robot requiring much user input, either to demonstrate the full task (GC) or to make many corrections due to inefficient learning (SA + GMR).
Furthermore, in the first teaching trial, which required the most correction, the maximum force applied by users was lower with LOPAL than with SA + P-GMR ($p=0.039$, $|r| = 0.83$, CI$_{95\%}$=[$0.17$, $1.00$]).
Since LOPAL slows down in areas with low local demonstration quality, users may no longer need to apply large initial forces when making corrections to manually slow the robot down. 
This could explain the tendency of LOPAL's subjective physical demand being rated lower, seen in Fig. \ref{fig:questionnaire} (\textbf{b}).
Additionally, participants reported lower mental demand with LOPAL than with the other SA-based teaching variants (SA + GMR and SA + P-GMR) ($p<0.027$, smallest $|r| = 0.90$, CI$_{95\%}$=[$0.35$, $0.99$]). 
This may be attributed to the active user feedback, which likely helped users to focus their corrections where they were most needed. \rev{Verbal reports from participants} provided insight on why the mental demand with GC resulted in similar ratings as with the SA-based teaching approaches: while shared autonomy reduced the mental effort associated with directly performing the task, it simultaneously introduced a new form of cognitive demand — monitoring and reacting to the robot’s autonomous behavior.

\subsubsection{Key findings}
Our experiment evaluation shows that in a real-world robotic task, the use of local quality information in LfD can enhance learning effectiveness, which can be further enhanced by bridging regions of low local performance through interpolation.

This efficient learning approach, which leverages local performance information, enabled a significant reduction in the physical teaching effort.
Furthermore, participant ratings indicate that providing feedback on missing high-quality data through speed reduction and visual cues has the potential to decrease the user’s mental load during incremental teaching.
Overall, these results suggest that our proposed LOPAL framework can increase learning efficiency from imperfect demonstrations while simultaneously reducing the teaching burden on the user.

\section{CONCLUSIONS}
This paper presents LOPAL -- a novel local performance-aware active learning from demonstration framework with two main contributions:
(1) A LfD approach that leverages locally varying performance information to generate a reference behavior that selectively replicates the best demonstrated local behaviors, thereby allowing it to outperform the original demonstrations.
(2) An active data collection method that facilitates incremental learning by prompting the user to correct the execution of this reference behavior in order to acquire additional informative samples.
Simulation and experiment results highlight the benefit of our proposed framework. LOPAL not only improved task performance but also reduced the demonstration effort for the user.
Future work could investigate the use of other quality measures for applying LOPAL to a wider range of tasks or extending the approach to larger state spaces, like dual-arm manipulation tasks.

\bibliographystyle{IEEE_custom}

\bibliography{ref}

\end{document}